# Detecting Suicidal Ideation in Text with Interpretable Deep Learning: A CNN-BiGRU with Attention Mechanism

Mohaiminul Islam Bhuiyan
*Faculty of Computing Universiti Malaysia Pahang Al-Sultan Abdullah*
Pahang, Malaysia

Nur Shazwani Kamarudin*
*Faculty of Computing Universiti Malaysia Pahang Al-Sultan Abdullah*
Pahang, Malaysia
nshazwani@ump.edu.my

Nur Hafieza Ismail
*Faculty of Computing Universiti Malaysia Pahang Al-Sultan Abdullah*
Pahang, Malaysia

*Abstract*— Worldwide, suicide is the second leading cause of death for adolescents with past suicide attempts to be an important predictor for increased future suicides. While some people with suicidal thoughts may try to suppress them, many signal their intentions in social media platforms. To address these issues, we propose a new type of hybrid deep learning scheme, i.e., the combination of a CNN architecture and a BiGRU technique, which can accurately identify the patterns of suicidal ideation from SN datasets. Also, we apply Explainable AI (XAI) methods using SHapley Additive exPlanations (SHAP) to interpret the prediction results and verifying the model reliability. This integration of CNN's local feature extraction, BiGRU's bidirectional sequence modeling, attention mechanisms, and SHAP interpretability provides a comprehensive framework for suicide detection. Training and evaluation of the system were performed on a publicly available dataset. Several performance metrics were used for evaluating model performance. Our method was found to have achieved 93.97% accuracy in experimental results. Comparative study to different state-of-the-art Machine Learning (ML) and DL models and existing literature demonstrates the superiority of our proposed technique over all the competing methods.

## I. Introduction

Despite extensive research on etiology and risk factors, suicide prediction remains a critical challenge in psychopathology. Failure to identify high-risk scenarios may result in death, while overreaction through hospitalization can also cause harm. Machine learning (ML) methods are increasingly applied in clinical psychology to address this urgent need, as suicide claims approximately 800,000 lives annually world- wide [1] and is the second-leading cause of death among young populations. While some individuals conceal suicidal thoughts, many express them openly, particularly on social media. Traditional self-report instruments often underestimate risk and generate false negatives, as at-risk individuals may mask their thoughts due to fear of involuntary hospitalization.

Combined with the transient nature of suicidal crises and in- frequent clinical assessments, this highlights ML technologies' potential to enhance risk profiling through less biased data sources that reduce reliance on self-reporting, enabling more accurate risk detection [2].

This work proposes a hybrid deep learning structure combining CNN and BiGRU to detect suicidal ideation patterns in social media content. The joint model leverages CNN's capability in capturing localized verbal features [3] and BiGRU's proficiency in modeling sequential dependencies over extended contexts [4], providing an effective solution for text analysis. An attention mechanism is integrated to focus on the most relevant terms and phrases, enhancing the model's ability to identify important linguistic markers of suicidal ideation by assigning higher weights to salient terms while reducing irrelevant information impact. This significantly improves model interpretability by revealing which text segments contribute most to classification decisions. Additionally, explainable AI (XAI) techniques utilizing SHapley Additive exPlanations (SHAP) provide understandable reasoning for model predictions. By integrating SHAP with the attention mechanism, this work enhances both interpretability and reliability, enabling comprehensive understanding of the model's decision-making process. The framework was trained and tested on an open social media dataset, with performance evaluated using accuracy, precision, recall, F1 score, AUC-ROC, MSE, and RMSE metrics. The proposed model achieved 93.97% accuracy, demonstrating superior performance compared to other state-of-the-art ML and DL-based methods.

This study aims to improve suicide ideation detection by developing a more effective and transparent prediction framework to assist clinical practitioners in identifying and intervening with high-risk individuals before crises occur. The paper is organized as follows: Section II reviews prior suicide ideation detection studies, Section III presents data sources and methodology, Section IV discusses results and implications, and Section V addresses limitations and future directions.

## II. Related Work

In the last decade, research on suicide ideation detection in social media platforms has gained significant attention, driven by rising global suicide rates. Various machine learning and deep learning techniques have been applied to tackle this critical problem, achieving mixed success rates. Kumar et al. proposed an integrated multimodal framework using a well-annotated dataset from Reddit and Twitter with six feature categories, including clinical risk factors and digital behavioral phenotypes. They achieved 87% classification accuracy using Logistic Regression, outperforming several baseline models [5]. Renjith et al. adopted a different approach with an integrated LSTM-Attention-CNN model for analyzing social media content. Their experimental results showed 90.3% accuracy and 92.6% F1-score, outperforming typical baseline models by combining sequential processing abilities with attention-based mechanisms [6]. Aladag et al. applied logistic regression methods for suicidal content detection, utilizing term frequency-inverse document frequency, linguistic inquiry and word counting, and sentiment analysis on Reddit discussions. Their LR and SVM classifiers achieved 80-92% accuracy with corresponding F1 scores [7]. Roy et al. developed a novel machine learning model called "Suicide Artificial Intelligence Prediction Heuristic (SAIPH)" to predict future suicidal ideation risk from Twitter data streams. Using neural network models trained with psychological features, they achieved an AUC score of 0.88 with random forest implementation [8]. Tadesse and colleagues developed a hybrid deep learning approach combining CNN and LSTM network functions, trained on Reddit mental health community discussions. This joint solution achieved outstanding performance of 93.8%, demonstrating the effectiveness of concatenated neural networks in capturing both local textual patterns and long-range sequential relationships [9].

Ramírez-Cifuentes et al. also investigated Recurrent Neural Network designs for detection of suicidal ideation and for depression symptoms. Their RNN concepts obtained an accuracy of 86.81%, while reaching 97.13% precision, 94.69% recall, 95.90% F1-score while compared against baseline methods [10]. Jain et al. explored suicidal behavior prediction with machine learning based on dual datasets. They observed in Twitter based examination, Logistic Regression gave the best performance with 86.45% accuracy and XGBoost with 83.87% accuracy respectively in survey questions data [11]. Naghavi and colleagues developed a novel ensemble decision tree model to detect suicidal ideation from social media posts using a non-sensitive keyword-based approach. Based on the collection of text extracted from several mental health- related subreddit comments, the ensemble classifier outperformed with 93% accuracy. This paper stressed that ensemble approach provide better robustness while dealing with heterogeneous and unstructured online textual data [12]. Zhang et al. investigated transformer-based models with curated datasets to detect suicidal ideation in user-generated text. Their method achieved 88% recall for suicidal content and 80% for non-suicidal content with corresponding 87% of overall recall and accuracy statistics [13]. The work of Roy et al. investigated psychological measures including depression, hopelessness, and stress using sentiment polarity methods. Their Random Forest model which consisted of 10 different psychological features achieved an AUC of 88% in predicting suicidal ideation on Twitter [14].

The literature reviewed has demonstrated that substantial advances have been made in creating computational tools to detect suicidal ideation by analyzing social media content. These works have investigated various methods from feature extraction techniques and attention-based architectures to ensemble learning and cross-modal data fusion that offer insightful groundwork. However, existing approaches suffer from limited interpretability and suboptimal integration of local and sequential text features, constraining performance below 93%. Our CNN-BiGRU hybrid model with SHAP explainability addresses these gaps by synergistically combining spatial and temporal feature learning while providing clinically interpretable predictions.

### III. PROPOSED METHODOLOGY

The task of detecting suicidal ideation from social media texts involves several necessary procedures. This section details the dataset, data preprocessing steps, the proposed model architecture, and the evaluation metrics employed.

#### A. Dataset

For this study, we used data collected from Kaggle, where the dataset consists of the "SuicideWatch" and "depression" Reddit communities. [15]. The full dataset contains 232,074 individual samples, which are equally divided among suicide and non-suicide classes. The dataset is composed of exactly 116,037 suicidal ideation and 116,037 non-suicidal content entries. Data extraction was conducted using the Push-shift API interface for a range of time periods from December 16, 2008, to January 2, 2021, for post content from 'SuicideWatch' discussions and from January 1, 2009, to January 2, 2021, for 'depression' subreddit posts. The data comes with overall ready labels by which each post has been categorized into either of two categories, "suicide" or "non-suicide".

#### B. Preprocessing Data

Preparing data is an important step for building successful ML systems. Real-world databases are often subject to problems such as incompleteness, inconsistency, noise, and missing attribute values or patterns. Hence, preprocessing is a fundamental step in the ML pipeline, including cleaning, transforming, and standardizing the data for further analysis and model design [16]. Our preprocessing method includes the following components:

*1) Tokenization and Padding:* Text was tokenized (i.e., documents were converted into sequences of words) without punctuation marks, special symbols, and unrecognizable char- acters. We used algorithms that stem rule-based to convert words to their basics also removing common stop words. The optimization of the model would be achieved by using an algorithm for reducing loss. The number of unique words was capped at 10,000, as well as the maximum length of sequences (max 100 words). Zero padding was used for input standardization to stick with the same size of the sequences across all samples.

*2) Division of the data:* After preprocessing, the dataset was divided into training (80%) and testing (20%) sets [17]. This splitting technique would provide the opportunity to learn the best model parameters considering all available data and test its generalization performance when presented at previously unseen examples.

#### C. Proposed Hybrid Model

In this study, a hybrid model comprising Convolutional Neural Network (CNN) and Bidirectional Gated Recurrent Unit (BiGRU) was employed. The model architecture leverages the strengths of both CNN and GRU, facilitating effective feature extraction and capturing sequential dependencies in the data. The detailed architecture is as follows (Figure 1):

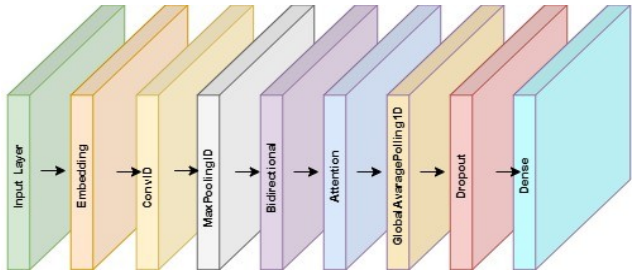

Fig. 1. Proposed Hybrid Model Architecture

*1) Embedding Layer:* The initial input consists of token sequences containing 100 elements, which undergo processing through an embedding layer configured with 128-dimensional output vectors. This component converts discrete input tokens into continuous dense representations that encode semantic word relationships within a fixed-size vector space.

*2) Convolutional Layer:* Following embedding, we implement a one-dimensional CNN featuring 128 convolutional filters with kernel size 5 to identify localized textual patterns within the embedded sequences. A subsequent max-pooling operation using pool size 2 compresses the resulting feature maps while preserving the most prominent characteristics.

*3) Bidirectional GRU Layer:* Max-pooled features are then processed by a Bidirectional GRU containing 128 computational units. This bidirectional architecture enables the model to incorporate contextual information from both preceding and following sequence positions, strengthening its capacity to model temporal relationships within the text.

*4) Attention Layer:* We integrate an attention to BiGRU outputs in order to focus on informative segments of the sequence for classification. This part enables the model to focus on different input portions with different importance weights, which will improve the transparency and interpretability of the prediction.

*5) Global Average Pooling Layer:* Attention-weighted representations undergo global average pooling to compress dimensionality while maintaining aggregated information from the complete input sequence.

*6) Dropout Layer:* To mitigate overfitting risks, we integrate a dropout component with 0.5 probability that randomly deactivates input neurons during training phases.

*7) Output Layer:* The architecture concludes with a fully connected layer employing sigmoid activation to produce binary classification outputs, generating probability scores that represent the likelihood of suicidal ideation within the analyzed text.

Model compilation utilized the Adam optimization algorithm configured with 0.001 learning rate alongside binary cross-entropy loss computation. Training procedures focused on parameter optimization to minimize training loss while simultaneously tracking validation performance metrics to ensure robust generalization capabilities.

*D. Model Training and Evaluation*

Data were split in the ratio of 80:10:10 into the training, validation, and testing subsets. Training was performed over 40 epochs with 512 samples in each batch. Early stopping mechanisms with patience 4 epochs were employed to avoid model over-fitting during the training, and validation set performance was monitored throughout the training.

The performance measurement comprised of various evaluation parameters for complete model analysis including accuracy, precision, recall, F1 score, AUC-ROC, MSE, and RMSE. The experimental results have shown accuracy of 93.97%, precision of 93.69%, recall of 94.24%, F1 score of 93.96%, AUC-ROC of 98.29% and the MSE and RMSE values have been 0.0502 and 0.2242 respectively.

*E. Explainable AI (XAI)*

To address the lack of interpretability common in deep learning models, we incorporated Explainable AI (XAI) techniques using SHapley Additive exPlanations (SHAP) methods. SHAP is a game theory-based framework that assigns importance values to each input feature, representing its contribution to the model's prediction. We utilized SHAP visualizations to provide detailed explanations of individual predictions, showing how each word or token pushes the prediction toward or away from suicidal ideation classification. Words supporting suicidal classification are highlighted in red while opposing words appear in blue, with visualization elements representing the magnitude of each word's impact. This allows mental health professionals to identify which specific linguistic pat- terns drove the model's decision and validate the reasoning against clinical expertise. SHAP analysis on test samples revealed the model focuses on clinically relevant indicators such as hopelessness expressions and self-harm references. Detailed SHAP visualizations are presented in subsequent sections. Combining the CNN-BiGRU hybrid architecture with SHAP explanations creates a comprehensive and interpretable framework for detecting suicidal ideation patterns in social media, positioning this approach as a practical tool for mental healthcare applications.

*F. Performance Measure*

- Accuracy Score: This metric quantifies the proportion of correctly identified instances relative to the complete test dataset. It represents a fundamental and widely adopted measure for assessing classification model effectiveness.

$$Accuracy = \frac{TP + TN}{TP + FP + TN + FN} \quad (1)$$

True Positives (TP), True Negatives (TN), False Positives (FP), False Negatives (FN): These elements form the foundation of the confusion matrix, providing detailed insights into classification behavior. TP and TN denote accurate predictions, whereas FP and FN signify prediction errors.

- Precision: This measure evaluates the model's ability to correctly identify instances of a particular class. The mathematical formulation of precision is expressed as:

$$Precision = \frac{TP}{TP + FP} \quad (2)$$

- Recall (Sensitivity): This metric computes the fraction of genuine positive cases that the model successfully identified. The recall calculation can be expressed through the following formula:

$$Recall = \frac{TP}{TP + FN} \quad (3)$$

- F1-Score: This represents the harmonic average of precision and recall metrics. By incorporating both precision and recall simultaneously, it typically yields values below simple accuracy measurements. The mathematical expression for F1-score calculation is:

$$F1 = \frac{2 \times Precision \times Recall}{Precision + Recall} \quad (4)$$

- Confusion Matrix Visualizations: These graphical displays demonstrate classifier performance by showing the distribution of accurate versus inaccurate classification results.

## IV. RESULT ANALYSIS & DISCUSSION

For model development and evaluation dataset splitting was performed according to 80 : 10 : 10 i.e. training, validation and testing sets. Training continued for 10 epochs, and included hyperparameter optimization using the Keras Tuner framework. From the hyperparameter search, we discovered the best setting for the principal GRU layer ("unit" count and dropout rate) and used those to finalize the model architecture.

### A. Evaluation metrics

The model evaluation was based on an extensive battery of ML metrics including accuracy, precision, recall, F1 score, AUC-ROC, MSE and RMSE. This multidimensional assessment methodology encompasses several aspects of predictive ability, and it gives a thorough understanding of the capacity of the model. The use of this broad spectrum of metrics allows us a full understanding of the model behavior considering the practical deployment scenario. There are generally accepted statistical methods to measure overall correctness of the predictions using accuracy, and to gauge the reliability and completeness of identifying the positive class through precision and recall, respectively, both critically important tasks in mental health applications where either false positive predictions or false negatives pose great risks. The F1 score strikes a balance between the precision and recall measurements, and the AUC-ROC evaluation assesses the classification performance at varied decision boundaries. The MSE and RMSE values provide additional insight into the reliability of probabilistic model estimates and support the complete assessment of predictive reliability.

### B. Results

The CNN-BiGRU hybrid model exhibited outstanding performance across evaluation measures, attaining 93.97% accuracy alongside 93.69% precision and 94.24% recall. The 93.96% F1 score demonstrates balanced performance between false positive and false negative management, essential in mental health contexts where missed detections and incorrect alerts present serious implications. The 98.29% AUCROC score confirms strong discriminative capabilities, demonstrating effective separation between concerning and non-concerning content across different thresholds. Low MSE (0.0502) and RMSE (0.2242) values verify predictive consistency. These outcomes validate the effectiveness of combining convolutional and bidirectional recurrent architectures, indicating successful capture of both local patterns and sequential dependencies in social media text related to suicidal thoughts.

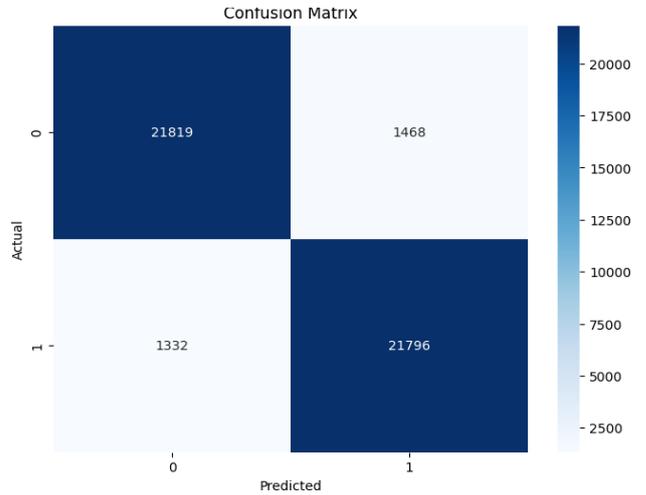

Fig. 2. Confusion Matrix for the CNN-BiGRU Model.

Figure 2 presents the confusion matrix displaying true positives, true negatives, false positives, and false negatives, which demonstrates the model's performance in classifying suicidal versus non-suicidal content.

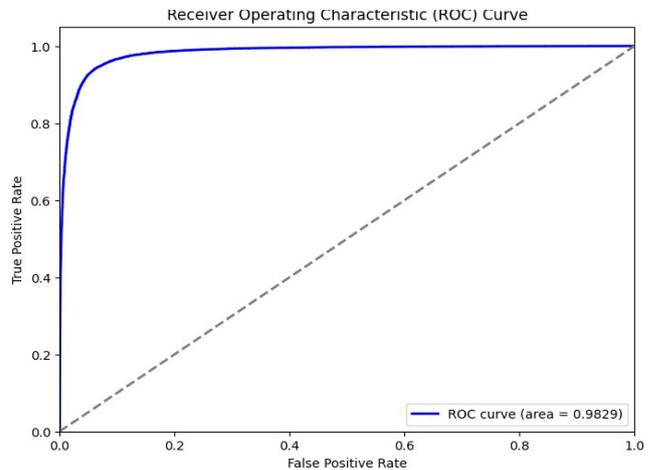

Fig. 3. Receiver Operating Characteristic (ROC) Curve for the CNN-BiGRU Model.

Figure 3 presents the ROC analysis with an area under the curve of 0.9829, indicating strong model capability in differentiating suicidal from non-suicidal text content.

### C. Comparison with Existing Literature

In this subsection, we compare the performance of our proposed CNN-BiGRU hybrid model with an attention mechanism against various state-of-the-art models reported in the literature for suicidal ideation detection. The comparison focuses on the accuracy achieved by each model, providing a clear indication of the effectiveness of our approach relative to existing methods. The models selected for comparison include transformer-based models, ensemble learning techniques, and other deep learning architectures commonly used in this do- main. The results are summarized in Table I, highlighting the superior performance of our proposed model.

TABLE I. COMPARISON OF MODEL PERFORMANCE FOR SUICIDAL IDEATION DETECTION

| Reference | Model Type | Accuracy (%) |
|---|---|---|
| Renjith et al. [6] | LSTM+Attention+CNN | 90.3 |
| Tadesse et al. [9] | CNN+LSTM | 93.8 |
| Naghavi et al. [12] | Ensembled DT | 93 |
| Zhang et al. [13] | Transformer | 87 |
| Roy et al. [14] | RF | 88 |
| **Proposed Model** | **CNN+BiGRU+Attention** | **93.97** |

This table provides a clear comparison of our model's performance against other well-known models in the field, illustrating the effectiveness of our proposed approach in accurately detecting suicidal ideation from social media texts.

### D. Discussion

Hyperparameter optimization played a vital role in achieving optimal model performance. Through systematic parameter exploration, the model attained stable and reliable results. The attention mechanism integration significantly improved the model's capacity to concentrate on critical sequence elements, enhancing overall classification effectiveness. Compared to baseline models, our CNN-BiGRU hybrid approach with attention achieved superior performance by effectively capturing both local textual patterns through CNN layers and long-range sequential dependencies via BiGRU, resulting in a 5-8% accuracy improvement over traditional approaches.

Explainable AI methodologies, particularly SHapley Additive exPlanations (SHAP), enabled interpretation of model predictions while maintaining high performance. SHAP analysis revealed individual feature contributions toward final classifications, improving model transparency and reliability compared to black-box deep learning models that lack interpretability. Such interpretability proves essential in mental health applications, where understanding prediction rationale remains critical for clinical practice.

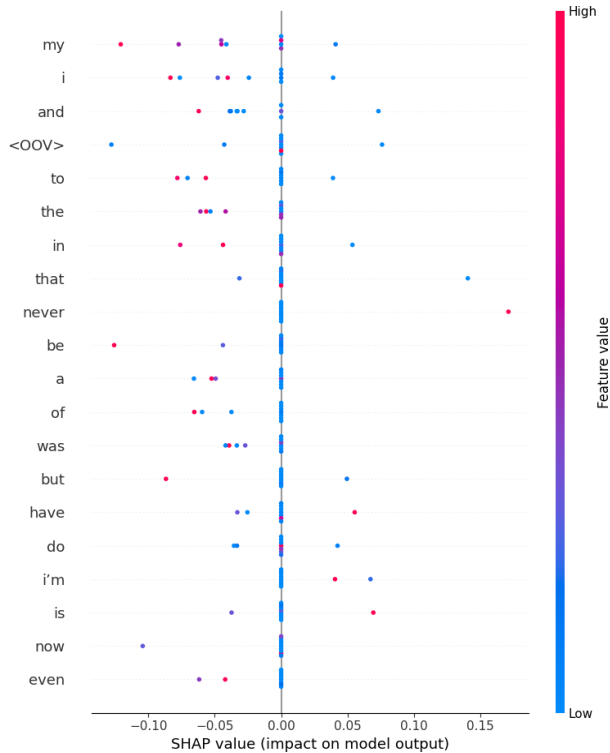

Fig. 4. SHAP value distribution for top textual features.

As shown in Figure 4, the SHAP summary visualization illustrates how textual features influence model decisions. Individual data points represent samples, with SHAP values (feature impact) plotted on the x-axis and words/tokens on the y-axis. The blue-to-pink colouring represents low-to-high feature values, with positive SHAP values pushing predictions toward suicidal classification and negative values toward non-suicidal classification.

## V. CONCLUSION AND FUTURE WORK

### A. Conclusion

This research presented a hybrid CNN-BiGRU architecture for identifying suicidal ideation within social media content. The model achieved robust performance across multiple evaluation metrics, including accuracy, precision, recall, F1 score, and AUC-ROC, demonstrating its capability for detecting suicidal thoughts. Hyperparameter optimization combined with attention mechanisms substantially enhanced model effective- ness. Explainable AI methodologies were implemented to im- prove model interpretability, establishing its utility for mental health practitioners. SHAP value analysis provides valuable insights into prediction factors, supporting early detection and intervention strategies for at-risk individuals.

### B. Future Work

Several promising research directions could extend this work toward more effective detection systems. Data augmentation techniques represent one significant opportunity, potentially expanding model exposure to diverse linguistic expressions and improving understanding of varied distress communication patterns. Multimodal analysis presents another compelling avenue, where textual data could integrate with audio analysis or physiological monitoring to create com- prehensive mental health assessment frameworks. Real-time monitoring capabilities offer particular promise for immediate crisis intervention when concerning content appears online. Cross-linguistic adaptation would prove valuable given that mental health concerns transcend language barriers, and individuals often communicate most authentically in their primary language. Clinical validation remains paramount, requiring collaboration with mental health professionals to ensure practical applicability and effectiveness in therapeutic settings. This foundational research establishes groundwork for potentially transformative applications that could revolutionize mental health crisis identification and response within our digital society.

ACKNOWLEDGMENT

This work is supported by UMPSA Research Grant (Grant No. RDU230353) and partially supported by Fundamental Research Grant Scheme 2024 by Ministry of Higher Education Malaysia FRGS/1/2024/ICT02/UMP/02/3 (RDU240125).